\documentclass{article}

\usepackage{spconfa4}
\usepackage{array}
\usepackage{booktabs}
\usepackage{cite}
\usepackage{graphicx}
\usepackage{amssymb,amsmath}
\usepackage{framed,multirow}
\usepackage{latexsym}
\usepackage{array}
\usepackage{balance}
\usepackage{booktabs}
\usepackage{caption}
\usepackage{color}
\usepackage{graphicx}
\usepackage{multirow}
\usepackage{times}
\usepackage{url}
\usepackage{threeparttable}
\usepackage{float}
\usepackage{multicol}

\usepackage{grffile}
\usepackage[numbers]{natbib}
\usepackage{rotating}
\definecolor{lightblue}{RGB}{173,216,230}
\definecolor{lightgreen}{RGB}{217,234,211}
\definecolor{lightyellow}{RGB}{255,246,218}
\definecolor{lightred}{RGB}{251,228,231}
\definecolor{atlantisblue}{RGB}{135,206,235} 
\definecolor{paleblue}{RGB}{234,250,249} 
\definecolor{palemint}{RGB}{246,251,243} 
\definecolor{palepeach}{RGB}{254,244,237} 
\definecolor{paleyellow}{RGB}{255,249,232} 
\definecolor{palepink}{RGB}{251,230,238} 
\definecolor{palered}{RGB}{255,227,227}
\definecolor{palepurple}{RGB}{238,233,246} 

\definecolor{phoenix}{RGB}{246,183,168} 
\definecolor{AGT}{RGB}{225,243,227} 
\definecolor{FSS}{RGB}{239,239,203} 
\definecolor{AIAIG}{RGB}{226,238,253}
\definecolor{MIIG}{RGB}{237,252,215} 
\definecolor{AIF}{RGB}{212,244,242}

\definecolor{theme}{RGB}{255,0,0} 
\definecolor{attri}{RGB}{208,125,60}
\definecolor{back}{RGB}{36,144,135} 
\definecolor{emo}{RGB}{112,48,160}

\newcommand{\p}[1]{\textcolor{black}{#1}}

\begin{document}

\title{
{Aesthetic Matters in Music Perception for Image Stylization: A Emotion-driven Music-to-Visual Manipulation}}

\name{Junjie Xu$^1$, Xingjiao Wu$^1$, Tanren Yao$^1$, Zihao Zhang$^1$, Jiayang Bei$^1$, Wu Wen$^1$, Liang He$^1$}
\address{
	$^1$East China Normal University, Shanghai, China~~~~~
}

\UseRawInputEncoding
\maketitle

\begin{abstract}
\p{Emotional information is essential for enhancing human-computer interaction and deepening image understanding. However, while deep learning has advanced image recognition, the intuitive understanding and precise control of emotional expression in images remain challenging. Similarly, music research largely focuses on theoretical aspects, with limited exploration of its emotional dimensions and their integration with visual arts. To address these gaps, we introduce EmoMV, an emotion-driven music-to-visual manipulation method that manipulates images based on musical emotions. EmoMV combines bottom-up processing of music elements—such as pitch and rhythm—with top-down application of these emotions to visual aspects like color and lighting. We evaluate EmoMV using a multi-scale framework that includes image quality metrics, aesthetic assessments, and EEG measurements to capture real-time emotional responses. Our results demonstrate that EmoMV effectively translates music’s emotional content into visually compelling images, advancing multimodal emotional integration and opening new avenues for creative industries and interactive technologies.}
\end{abstract}


\section{Introduction}
% Images contain multi-level information, including low-level features (such as color and texture), mid-level features (such as shape and edges), high-level semantic objects (such as people and scenes), and emotional and aesthetic information \cite{joshi2011aesthetics}. In particular, the emotional information within images plays a critical role in enhancing human-computer interaction experiences, promoting the development of creative industries, and achieving a deeper understanding of images. However, despite the significant progress made by current deep learning approaches in image recognition, classification, and other sub-tasks, there remains a lack of systematic exploration of emotional information. Specifically, the emotional expression in images is difficult to understand intuitively and precisely control, making the generation of emotionally expressive images a long-standing challenge. Therefore, \textbf{effectively controlling and generating images with emotional expression remains an urgent research problem}.
Images contain multi-level information, including low-level features (e.g., color, texture), mid-level features (e.g., shape, edges), high-level objects (e.g., people, scenes), and emotional and aesthetic content \cite{joshi2011aesthetics}. Emotional information is crucial for improving human-computer interaction, advancing creative industries, and deepening image understanding. However, despite advancements in deep learning for image recognition and classification, emotional expression in images remains underexplored. Specifically, \textbf{intuitively understanding and precisely controlling emotional expression in images is challenging}, making it a persistent research problem.

% \begin{figure}[!t]
%     \centering
%     \includegraphics[width=1\linewidth]{figs/intro_v1.pdf}
%     \caption{Demonstration }
%     \label{fig:intro}
% \end{figure}
% Music, as a powerful medium for emotional expression, can intuitively evoke human emotional resonance. Understanding the emotional expression within music not only aids in a deeper comprehension of the music itself but also holds significant importance for higher-level cognitive understanding. Currently, research in the field of deep learning applied to music primarily focuses on the extraction of theoretical information, such as instrument recognition and beat analysis, while the exploration and representation of emotional information are relatively limited. Although some studies have attempted to analyze music emotions through emotion-annotated datasets and emotion classification models \cite{yoo2024emotion}, these methods mainly rely on textual annotations and fail to fully capture the fine-grained and multi-dimensional characteristics of musical emotional expression. Consequently, \textbf{the richness and complexity of musical emotions are difficult to encapsulate through simple textual descriptions comprehensively}.

Music, a powerful medium for emotional expression, can evoke human emotional resonance. Understanding its emotional content not only deepens music comprehension but also enhances higher-level cognitive understanding. Current research on music mainly focuses on theoretical tasks like instrument recognition and beat analysis, with limited exploration of emotional expression. While some studies use emotion-annotated datasets and classification models \cite{yoo2024emotion}, they rely on textual annotations and fail to capture the fine-grained, multi-dimensional nature of musical emotions. Therefore, \textbf{the richness and complexity of musical emotions cannot be fully encapsulated by simple textual descriptions}.
% Meanwhile, numerous creative projects \cite{clemente2022musical, djalalova2023piano, belfi2018rapid} integrate music and visual arts to evoke human emotions, and studies \cite{talamini2022musical, boltz2009audiovisual, ebendorf2007impact} have demonstrated that the synergy between music and imagery fosters a deeper emotional comprehension.
% Integrating visual imagery can facilitate the extraction of emotional information from music, while simultaneously allowing the emotional cues derived from music to enhance the emotional dimensions of images. This interdisciplinary fusion not only improves the precision and richness of emotional expression in visual representations but also introduces innovative approaches and methodologies for emotion-driven image generation. In light of this, adopting a multimodal research approach to construct an association model between music and images is meaningful. Several studies \cite{sung2023sound, lee2020crossing, wan2019towards, lee2022sound} have explored cross-modal understanding between auditory information and visual images. However, these works predominantly emphasize the extraction of semantic information from audio and the generation of cross-modal representations, often overlooking the emotional dimension. Consequently, effectively \textbf{correlating the emotional information inherent in music and visual arts remains a highly challenging research topic}.
Many creative studies \cite{clemente2022musical, djalalova2023piano, belfi2018rapid} combine music and visual arts to evoke emotions, and studies \cite{talamini2022musical, boltz2009audiovisual} show that their synergy deepens emotional understanding. Integrating visual imagery can help extract emotional information from music, while music’s emotional cues can enhance the emotional depth of images. This interdisciplinary fusion improves emotional expression in visuals and introduces new methods for emotion-driven image generation. Thus, a multimodal approach to linking music and images is valuable. While studies \cite{sung2023sound, lee2020crossing, wan2019towards, lee2022sound} have explored cross-modal relationships between audio and visual information, they often focus on semantic extraction, neglecting the emotional dimension. Therefore, \textbf{correlating emotional information in music and visuals remains a significant challenge}.

Therefore, we proposed an \textbf{Emo}tion-driven \textbf{M}usic-to-\textbf{V}isual manipulation method, termed \textbf{EmoMV}, to manipulate an image with the condition from music.
Inspired by human cognitive processes \cite{gregory1970intelligent}, we adopt a strategy that seamlessly integrates bottom-up (music-to-emotion) and top-down (emotion-to-image) stages to translate emotional information from music into visual imagery.
% Specifically, learning from \cite{huron2008sweet, krumhansl1997exploratory} which state that emotional responses are elicited through the basic structural elements of music (such as pitch, rhythm, and chord changes), we start from the most fundamental theoretical information, with the supplimentary information from visual arts, and gradually map it to higher-level emotional dimensions. Meanwhile, for images, a top-down cognitive strategy can be employed to gradually map perceived emotional information to the low-level features of images, ultimately reflecting the overall aesthetic effect of the image. This approach not only helps in more accurately capturing and expressing emotional information but also enhances the emotional richness and expressiveness of cross-modal generated images.
Building upon the research of \cite{huron2008sweet, krumhansl1997exploratory}, which demonstrates that emotional responses are elicited by the fundamental structural elements of music—such as pitch, rhythm, and chord changes—we begin with these basic theoretical components. By incorporating supplementary information from visual arts, we progressively map these foundational elements to higher-level emotional expression (up-bottom $\downarrow$).
While, learning from the work \cite{xiao2024atlantis} which validates aesthetic attributes can be related to emotional expression, we further embody emotions into the aesthetic-related low-level dimensions like light, exposure, and color (bottom-up $\uparrow$).

% Another key challenge lies in evaluating emotion-driven manipulations, due to the inherent subjectivity of both aesthetic and emotional experiences. To address this, we propose a multi-scale evaluation approach that combines traditional image quality assessments, aesthetic evaluations, and electroencephalogram (EEG) measurements.
% EEG allows us to measure real-time emotional responses to the generated images, offering a more objective and immersive understanding of how well the images capture the emotional content of the music. This dual evaluation approach provides a comprehensive measure of the effectiveness of our method in translating music’s emotional qualities into visual output.
A key challenge in evaluating emotion-driven manipulations is the subjectivity of both aesthetic and emotional experiences. To address this, we propose a multi-scale evaluation approach combining image quality assessments, aesthetic evaluations, and EEG measurements. EEG tracks real-time emotional responses to images, providing a more objective and immersive understanding of how well the images capture the music’s emotional content. This dual evaluation offers a comprehensive measure of our method’s effectiveness in translating music’s emotional qualities into visual output.

The Contributions of this paper are as follows:
\begin{itemize}
    \item We develop a two-stage framework called EmoMV, which enables emotion-driven image manipulation. EmoMV leverages bottom-up and top-down cognitive strategies to construct emotional representations of music and images, in line with cognitive science principles.
    \item We introduce a Mus-Vis Textual Alignment module for aligning musical description and visual description with emotional expression. While Emotion-aware Aesthetic Image Refinement is proposed for mapping emotional information into visual elements like lighting, and exposure, and achieving the harmony of the whole image. 
    \item We propose a multi-scale evaluation framework and demonstrate the effectiveness of EmoMV through extensive experiments on a 38k music-image pair dataset that we collected from online sources. Our results show that EmoMV can positively impact emotional well-being, particularly in applications such as art therapy.
\end{itemize}

\section{Related Work}
\textbf{Image Emotion.}
% In recent years, the expression of emotional content in images has been extensively studied, with visual features designed and extracted at different levels. Yanulevskaya \cite{yanulevskaya2008emotional} first proposed a method for emotional classification of artworks based on low-level features. The impact of composition, color contrast, and texture features on image emotional expression is discussed in \cite{machajdik2010affective}, with these features designed according to artistic principles to capture emotional information in images. Traditional methods failed to encompass all the important factors related to human emotions, resulting in unsatisfactory outcomes. Recent attempts typically focus on extracting semantic features of the overall image, often neglecting the importance of higher-level abstract features. Yang \cite{yang2018visual} utilize object detection and attention mechanisms \cite{yang2018weakly} to assist emotion recognition. With specially designed emotional features, Yang \cite{yang2021stimuli} constructed a network aimed at learning emotions from different visual stimuli and mining the correlations between them \cite{yang2021solver}. However, given the abstract nature of emotion, development remains unsatisfactory, making it necessary to introduce auxiliary information to assist visual emotion recognition.
Recent studies have explored emotional content in images by designing and extracting visual features at different levels. Yanulevskaya \cite{yanulevskaya2008emotional} proposed an emotional classification method for artworks based on low-level features. The influence of composition, color contrast, and texture on emotional expression is discussed in \cite{machajdik2010affective}, where features are designed according to artistic principles. Traditional methods, however, failed to capture all factors influencing human emotions. More recent approaches focus on extracting semantic features but often overlook higher-level abstract elements. Yang \cite{yang2018visual} used object detection and attention mechanisms for emotion recognition, while \cite{yang2021stimuli} introduced a network to learn emotional correlations from visual stimuli. Despite these efforts, the abstract nature of emotion still poses challenges, requiring the integration of auxiliary information for more accurate visual emotion recognition. 
\\
\textbf{Music Emotion.}
% Music emotion research \cite{agrawal2021transformer}, as an interdisciplinary field spanning music theory, cognitive science, and artificial intelligence, has gained significant attention in recent years. Scholars have explored various aspects of how music conveys and evokes complex emotional experiences, from emotion recognition \cite{sarkar2020recognition} and classification \cite{panda2020audio} to generation \cite{copet2024simple, lam2024efficient, schneider2024mousai}. One of the core challenges in music emotion research is how to effectively represent emotions. Traditional approaches typically employ discrete emotion models, such as recognizing specific emotions like happiness, sadness or anger. However, Chaturvedi \cite{chaturvedi2022music} emphasizes the multidimensional nature of emotional experiences and advocates for more detailed frameworks than simple textual annotations. Relying solely on subjective listener feedback or textual descriptions for annotation methods fails to fully capture the complexity of musical emotions.
Music emotion research \cite{agrawal2021transformer}, an interdisciplinary field combining music theory, cognitive science, and AI, has gained significant attention. Researchers have explored emotion recognition \cite{sarkar2020recognition}, classification \cite{panda2020audio}, and generation \cite{lam2024efficient, schneider2024mousai}. A key challenge is effectively representing emotions. Traditional methods use discrete emotion models, identifying emotions like happiness or sadness. However, Chaturvedi \cite{chaturvedi2022music} highlights the multidimensional nature of emotional experiences, advocating for more nuanced frameworks beyond simple annotations. Relying on subjective feedback or textual descriptions alone fails to capture the full complexity of musical emotions.
\\
\textbf{Music-image Alignment.}
The alignment of music and visual \cite{chen2021CVPR, wu2022wav2clip, guzhov2022audioclip} to evoke and enhance emotional responses has been a key focus in the field of deep learning. Music has the ability to evoke complex emotional states, and when combined with visual, it can provide a richer emotional context \cite{talamini2022musical, boltz2009audiovisual, ebendorf2007impact}. This combination not only stimulates the emotional centers of the brain but also enhances the expressiveness of visual content. 
% Although recent studies have explored the fusion of music and images, most of them \cite{sung2023sound, lee2020crossing, wan2019towards, lee2022sound} focus primarily on semantic content, often overlooking the emotional dimension, which is crucial for achieving truly meaningful multi-modal experiences. 
% However, effectively correlating emotional information between these two modalities remains a significant challenge. Unlike semantic alignment, emotional alignment requires capturing the subtle emotional nuances in both modalities \cite{braun2019depression, lv2024auditory}, rather than explicit features. Meanwhile, the complexity of emotions varies across different music genres \cite{jostrup2023effects} and visual styles.
% The alignment of music and visuals \cite{chen2021CVPR, mazumder2021CVPR, wu2022wav2clip, guzhov2022audioclip} to enhance emotional responses has been a major focus in deep learning. Music evokes complex emotions, and when paired with visuals, it provides a richer emotional context \cite{talamini2022musical, boltz2009audiovisual, ebendorf2007impact}.
While recent studies explore music-image fusion, most \cite{sung2023sound, lee2020crossing, wan2019towards, lee2022sound} focus on semantic content, often neglecting the emotional dimension. 
Effectively correlating emotional information across these modalities remains challenging, as emotional alignment requires capturing subtle nuances, not just explicit features \cite{braun2019depression, lv2024auditory}. Additionally, emotions vary across music genres \cite{jostrup2023effects} and visual styles.
\\
\textbf{EEG-based Emotional Perception.}
% Emotional perception in the audio-visual modality has become a key area of research. Visual and auditory stimuli can each evoke distinct emotional responses. For example, visual stimuli such as facial expressions and emotional images effectively trigger emotions like happiness, sadness, or fear \cite{lang1997international}, while auditory stimuli, such as music and vocal tone, influence emotional perception \cite{juslin2003communication}. When visual and auditory cues are congruent, emotional perception tends to be consistent; however, when there is a mismatch between these cues, emotional responses may fluctuate or become ambiguous \cite{muller2012crossmodal}. Electroencephalography (EEG) has emerged as a valuable tool for detecting emotional fluctuations induced by audiovisual stimuli, especially in tracking changes in brain activity across different emotional states \cite{kamble2023comprehensive}. Unlike emotion recognition methods relying on subjective interpretation, EEG provides an objective measure of emotional responses, offering an accurate assessment of emotional shifts, particularly in complex multimodal environments \cite{zheng2015investigating}.
Emotional perception in the audio-visual modality has become a key research focus. Visual and auditory stimuli each evoke distinct emotions, with facial expressions and emotional images triggering feelings like happiness or fear \cite{lang1997international}, while music and vocal tone influence emotional perception \cite{juslin2003communication}. When these cues align, emotional perception is consistent; however, mismatched cues can lead to fluctuating or ambiguous responses \cite{muller2012crossmodal}. EEG has become a valuable tool for tracking emotional fluctuations induced by audiovisual stimuli, providing an objective measure of emotional shifts, particularly in complex multimodal contexts \cite{kamble2023comprehensive}. Therefore, we propose an EEG-based Multi-scale evaluation framework for evaluating our method.

\section{Methodology}
\begin{figure*}
    \centering
    \includegraphics[width=1\linewidth]{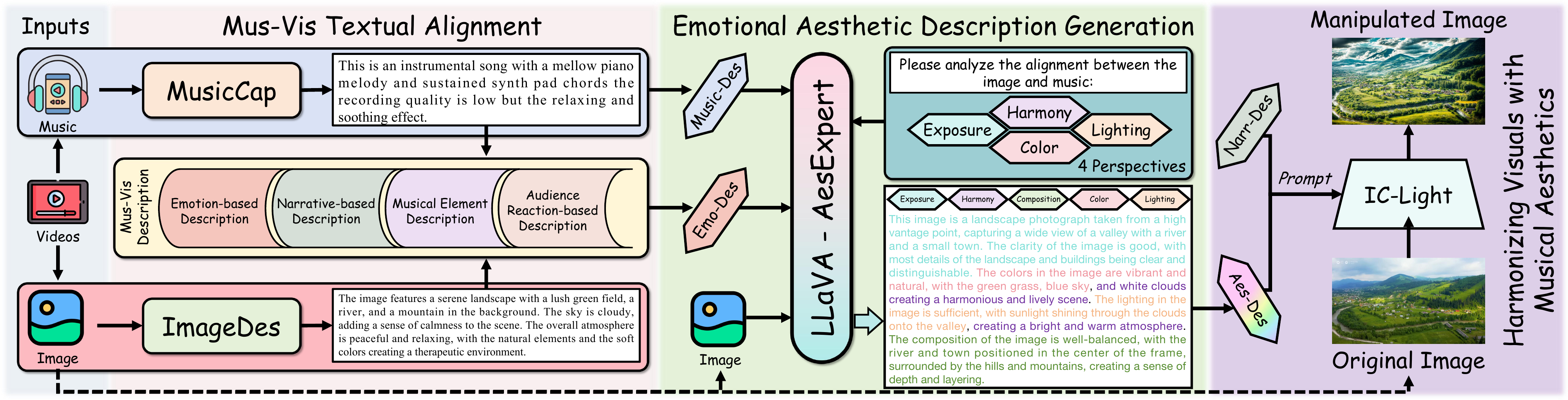}
    \caption{Overview of our framework Emotion-driven Music-to-Visual manipulation method, termed as \textbf{EmoMV}. \textbf{EmoMV} is composed of \textit{Mus-Vis Textual Alignment}, \textit{Emotional Aesthetic Description via Lighting}, and \textit{Harmonizing Visuals with Musical Aesthetics}, facilitating the extraction of emotional information and exhibit on the generated image.}
    \label{fig:framework}
\end{figure*}
In this section, we detail EmoMV, a two-stage approach comprising a \textit{Mus-Vis Textual Alignment} and a \textit{Emotion-aware Aesthetic Image Refinement}. The former correlates the musical elements with emotional expression, while the latter externalizes emotional information into the aesthetic attributes of the image.

\subsection{Task Definition}
% Given a music $M$ and an image $I$, what we need is to design a method $F$ that generates the image $I_{final}$ emotional related to $M$ and maintaining the semantic information from $I$: $I_{final} = F(I, M)$.
Given a music piece $M$ and an image $I$, the goal is to design a method $F$ that generates an emotionally relevant image $I_{final}$ while preserving the semantic content of $I$: $I_{final} = F(I, M)$.

\subsection{Mus-Vis Textual Alignment (MVTA)}
% 为了从音乐中获取情感的表达，我们将对于音乐基础元素的理解上升到情感的表达。其中，我们提出一种结合图片辅助的情感对齐方法，结合具体的被编辑图片本身的信息，来增强音乐对于这张图的情感表达。我们首先对于音乐基本元素进行提取，我们从audiocaptioning工作中获取灵感，提出利用编码器解码器结构进行获取音乐元素描述，利用强大的音频编码器HTSAT进行音乐编码，并利用BART进行解码，将整个模型在MusicCaps和MusicBench两个数据集上进行训练。在获取到音乐元素描述后，我们结合图片输入，利用强大的多模态模型LLaVA进行四维度描述的生成（Emotional, Narrative, Musical, Audience-reaction），将音乐基本元素描述结合图片映射到情感相关的描述中。
To extract emotional expression from music, we elevate our understanding of the fundamental elements of music to the level of emotional expression. In this process, we propose a novel emotion alignment method that incorporates image-based assistance, leveraging the specific information of the image being edited to enhance the emotional expression of music in relation to the image.
We begin by extracting the basic elements of music, drawing inspiration from \cite{mei2024wavcaps}. We introduce an encoder-decoder framework to generate musical element descriptions, using the powerful audio encoder HT-SAT \cite{chen2022hts} for music encoding and BART for decoding. The model is trained on the MusicCaps \cite{agostinelli2023musiclm} and MusicBench \cite{melechovsky-etal-2024-mustango} datasets. Once the musical element descriptions are obtained, we integrate them with the image input, employing the robust multimodal model LLaVA to generate four-dimensional descriptions (emotional, narrative, musical, and audience reaction). This allows us to map the music element descriptions and the image input into emotion-related descriptions:
\begin{equation}
    Description=LLaVA\{[HT-SAT + BART](M), I\}
\end{equation}

\subsection{Emotion-aware Aesthetic Image Refinement (EAIR)}
% 图像整体的情感表达来源于像素级的基础改变，我们将音乐中抽象的情感表达映射到基础的美学属性元素中，自上而下地将音乐情感切实落入到图像的基础元素中。这部分分为两个子部分：Emotional Aesthetic Description via Lighting：利用音乐描述、情感描述、图片，让AesExpert微调的模型将情感描述落入到图像美学的属性中进行体现，并获取相应情感展现图片中各个属性的描述；Harmonizing Visuals with Musical Aesthetics：利用IC-light模型结合我们的图像情感属性描述进行图像编辑，并加入的Narrative描述进行语义补全，保持编辑图像前后语义一致。
The overall emotional expression of an image originates from pixel-level changes. We map the abstract emotional expression in music to fundamental aesthetic attributes, applying a top-down approach to translate musical emotions into the core elements of the image. This process consists of two main parts:\\
\textbf{Emotional Aesthetic Description Generation:} By utilizing music descriptions, emotional annotations, and the image itself, we fine-tune the AesExpert model to embed emotional descriptions into the aesthetic attributes of the image. This step allows us to capture how emotional expressions manifest through various visual attributes of the image.\\
\begin{equation}
    Aes-Des = AesExpert(I, Emo-Des, Music-des)
\end{equation}
\textbf{Harmonizing Visuals with Musical Aesthetics:} We use the IC-Light model, along with our image emotion attribute descriptions, to edit the image. Additionally, we incorporate Narrative descriptions for semantic completion, ensuring semantic consistency before and after the image editing process.
\begin{equation}
    I_{final} = IC-Light(I, Aes-Des, Narr-Des)
\end{equation}

\begin{figure*}[h!]
    \centering
    \includegraphics[width=0.95\linewidth]{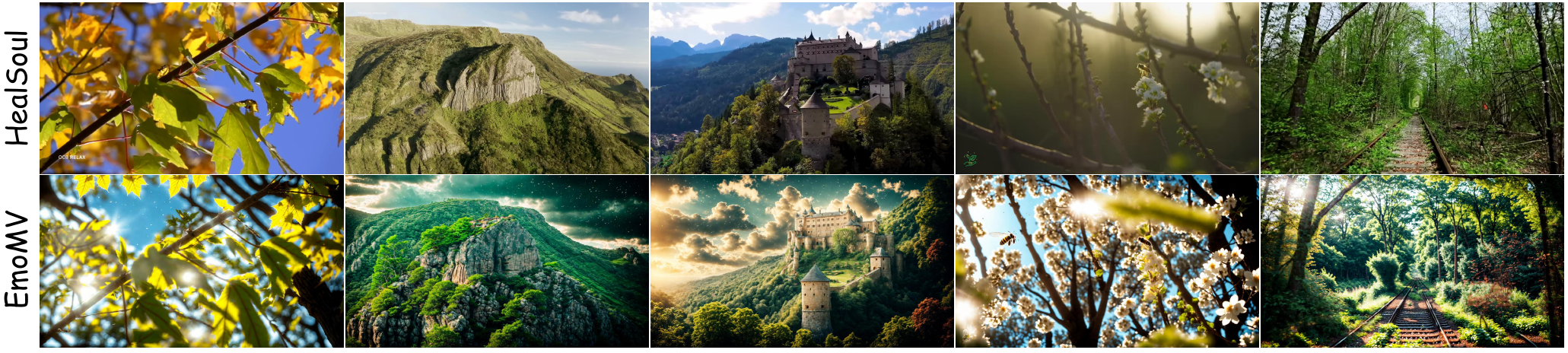}
    \caption{A qualitative analysis is conducted by contrasting the original image from HealSoul and the generated image through EmoMV. With the paired healing music, EmoMV generates images with more light, better composition.}
    \label{fig:qual}
\end{figure*}

\begin{table}[htbp!]
\tabcolsep=0.005cm
\centering
\caption{Comparison of Image Quality Metrics for \textit{HealSoul-1k} and \textit{EmoMV-1k} Datasets. Better results are highlighted in \colorbox{lightred}{pink}.}
\begin{tabular}{llcc}
\hline
\multicolumn{2}{l}{Metric}                                          & HealSoul-1k                                                & EmoMV-1k                                                  \\ \hline
\multicolumn{1}{l|}{\multirow{2}{*}{Sharpness}}    & (Mean ± Std) ↑ & $354.48 \pm 440.19$                                        & \colorbox{lightred}{$656.58 \pm 376.85$} \\
\multicolumn{1}{l|}{}                              & (Min, Max)     & $(1.63, 4292.79)$                                          & $(3.42, 1900.68)$                                           \\ \hline
\multicolumn{1}{l|}{\multirow{2}{*}{Contrast}}     & (Mean ± Std) ↑ & $0.97 \pm 0.07$                                            & \colorbox{lightred}{$0.99 \pm 0.04$}     \\
\multicolumn{1}{l|}{}                              & (Min, Max)     & $(0.40, 1.00)$                                             & $(0.36, 1.00)$                                              \\ \hline
\multicolumn{1}{l|}{\multirow{2}{*}{Colorfulness}} & (Mean ± Std) ↑ & \colorbox{lightred}{$143.74 \pm 29.33$} & $142.73 \pm 22.47$                                          \\
\multicolumn{1}{l|}{}                              & (Min, Max)     & $(35.48, 186.77)$                                          & $(60.94, 184.75)$                                           \\ \hline
\multicolumn{1}{l|}{\multirow{2}{*}{BRISQUE}}      & (Mean ± Std) ↓ & $37.13 \pm 16.75$                                          & \colorbox{lightred}{$28.07 \pm 13.31$}   \\
\multicolumn{1}{l|}{}                              & (Min, Max)     & $(1.94, 106.20)$                                           & $(-0.97, 71.91)$                                            \\ \hline
\end{tabular}
\label{tab:quality}
\end{table}

\begin{table*}[htbp!]
\tabcolsep=0.72cm
\caption{Aesthetic comparison of images derived from VEGAS ~\cite{zhou2018visual}, VGGSound ~\cite{chen2020vggsound}, HealSoul-1k, and EmoMV-1k. The best results are highlighted in \colorbox{lightred}{pink}, while suboptimal results are marked in \colorbox{lightgreen}{green}.}
\begin{tabular}{@{}cccccc@{}}
\toprule
\multicolumn{2}{c}{Datasets}                          & VEGAS\cite{zhou2018visual}  & VGGSound\cite{chen2020vggsound} & HealSoul-1k & EmoMV-1k \\ \midrule
\multicolumn{1}{c|}{\multirow{2}{*}{VILA}} & mean $\uparrow$     & 0.1891 & 0.3232   & \colorbox{lightgreen}{0.5694}      & \colorbox{lightred}{0.6304} \\
\multicolumn{1}{c|}{}                      & variance $\downarrow$ & \colorbox{lightred}{0.0042} & 0.0152   & 0.0138    & \colorbox{lightgreen}{0.0084} \\ \bottomrule
\end{tabular}
\label{datasets_vila}
\end{table*}

\section{Experiments}

\subsection{Datasets \& Experiment Settings}
We collected a dataset of 3.8k music-image pairs by searching for healing-related keywords on online resource and conducted experiments on this dataset to demonstrate the effectiveness of our method. We propose a multi-dimensional evaluation framework that includes metrics from the perspectives of image aesthetics, image quality, and human feedback. Specifically, we assess the following aspects:\\
\textbf{VILA Score:} 
In image synthesis, evaluating aesthetic quality is critical. To reduce subjective bias, we adopt the VILA framework~\cite{ke2023vila}, an advanced method for objective and reliable aesthetic assessment.
\\
\textbf{Image Quality Assessment:}
We assess image quality using four key metrics: \underline{Sharpness}, which quantifies edge clarity and detail; \underline{Contrast}, evaluating luminance and color differences to enhance visual depth; \underline{Colorfulness}, measuring the richness and vibrancy of hues; and \underline{BRISQUE} \cite{mittal2012making}, a no-reference metric that evaluates perceptual quality using natural scene statistics. These metrics collectively provide a comprehensive evaluation of visual quality.
\\
\textbf{Prefrontal EEG Analysis:}
We analyze prefrontal EEG signals to assess the emotional and cognitive impact of generated images. The detected brainwave activity is closely linked to users’ emotional states and cognitive engagement, offering real-time, objective insights that complement traditional quality assessments.

\subsection{Qualitative Analysis}
% We conducted a qualitative analysis to demonstrate how the EmoMV method impacts the emotional and aesthetic qualities of generated images. We selected 10 images from the \textit{HealSoul} dataset and compared them with \textit{EmoMV}-generated images using the corresponding music input.
% As demonstrated in Fig. \ref{fig:qual}, the HealSoul dataset, consisting of 3.8k natural landscape images, conveys a bright, warm aesthetic that evokes a soothing, therapeutic atmosphere. These images are visually appealing, with vibrant lighting and harmonious compositions. In comparison, images generated with the EmoMV method, incorporating emotional cues from music, feature brighter lighting, enhanced stylistic expression, and improved visual EmoMV, resulting in more vivid, emotionally resonant visuals that offer a richer and more engaging aesthetic experience.
We conducted a qualitative analysis to assess the impact of \textit{EmoMV} on the emotional and aesthetic qualities of generated images. We selected 10 images from the \textit{HealSoul} dataset and compared them with EmoMV-generated images using corresponding music inputs. As shown in Fig. \ref{fig:qual}, the HealSoul dataset, consisting of 38k natural landscape images, conveys a bright, warm aesthetic that evokes a soothing, therapeutic atmosphere. In contrast, EmoMV-generated images, enhanced with emotional cues from music, exhibit brighter lighting, richer stylistic expression, and improved visual EmoMV, resulting in more vivid, emotionally resonant visuals.

\begin{figure*}
    \centering
    \includegraphics[width=1\linewidth]{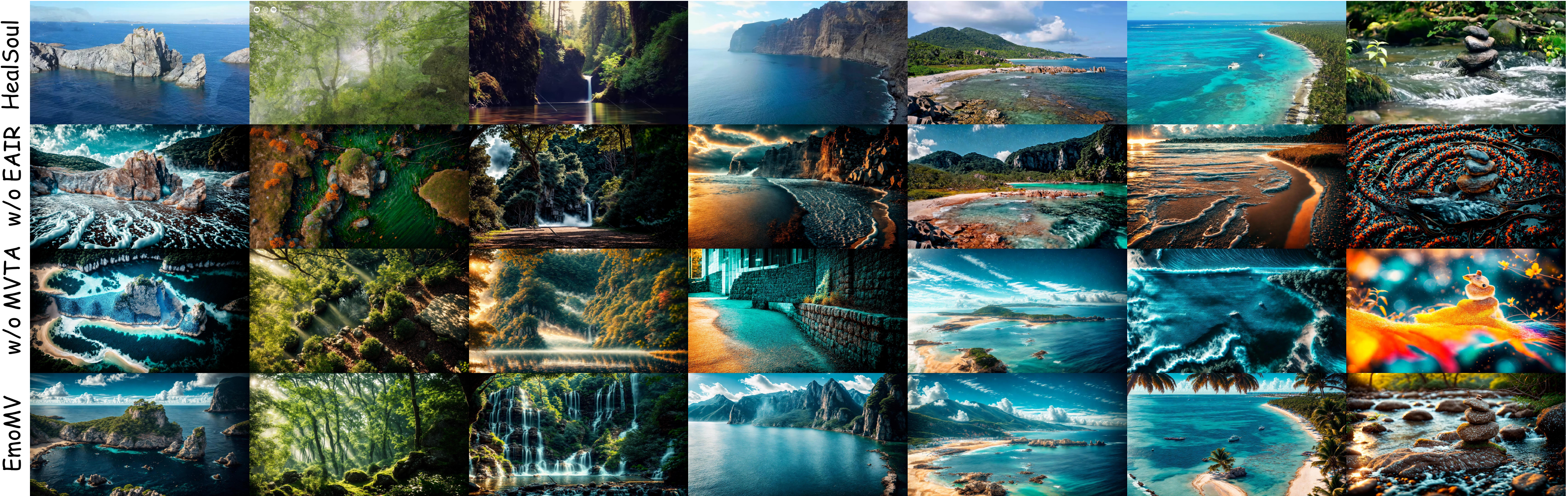}
    \caption{Ablation study for the effect of prompt input into the iclight, we put images from \textit{HealSoul} and generated from EmoMV branches without EAIR (w/o EAIR) or without MVTA (w/o MVTA).}
    \label{fig:ablation}
\end{figure*}

\begin{table}[htbp!]
\tabcolsep=0.35cm
\caption{EEG signal comparison between the \textit{HealSoul-1k} and \textit{EmoMV-1k} datasets. Superior results are highlighted in \colorbox{lightred}{pink}.}
\begin{tabular}{llcc}
\hline
\multicolumn{2}{l}{Frequency bands}                    &\textit{HealSoul-1k} & \textit{EmoMV-1k}   \\ \hline
\multicolumn{1}{l|}{\multirow{2}{*}{Alpha $\downarrow$}} & Mean   & $27306.10$   & \colorbox{lightred}{$27905.21$} \\
\multicolumn{1}{l|}{}                       & Events & $5$         & \colorbox{lightred}{$4$}        \\ \hline
\multicolumn{1}{l|}{\multirow{2}{*}{Beta $\uparrow$}}  & Mean   & $19211.42$  & \colorbox{lightred}{$27996.47$} \\
\multicolumn{1}{l|}{}                       & Events & $8$         & \colorbox{lightred}{$21$}       \\ \hline
\multicolumn{1}{l|}{\multirow{2}{*}{Gamma $\uparrow$}} & Mean   & $7691.83$  & \colorbox{lightred}{$9028.21$} \\
\multicolumn{1}{l|}{}                       & Events & $5$         & \colorbox{lightred}{$8$}        \\ \hline
\end{tabular}
\label{tab:eeg}
\end{table}

\subsection{Quantitative Analysis}
% To quantitatively assess the performance of the EmoMV method, we conducted a series of experiments focusing on both the visual quality and emotional alignment of the generated images. We applied multiple evaluation metrics to measure the impact of music-driven emotional cues on image aesthetics and emotional resonance. Specifically, we analyzed the image quality using state-of-the-art aesthetic models and compared the emotional consistency between the input music and the generated images. Additionally, we employed EEG measurements to capture user responses and evaluate the emotional engagement with the images, providing a comprehensive assessment of the effectiveness of our approach. The following sections present the results of these quantitative evaluations.
% 我们设计了对于图片质量评估（客观），美学大模型评估（半客观半主观），EEG监测（主观）的多尺度评价体系进行评估，如表1，2，可见HealSoul数据集具有较高的美学质量，EmoMV所生成的图片相较于HealSoul具有更高的图像质量、美学表现，通过音乐情感引导对于图像的质量和美感进行了提高。我们随机选取了15个人进行了随机100张图片-音乐的试听同步呈现，监测EEG脑波表现，如表3所示，EmoMV的XX波表现更好，表现出更好的情感表现，具有契合的视听一致性，证明了EmoMV能够编辑生成情感维度一致的图片的能力。
We designed a multi-scale evaluation system that includes image quality assessment (objective), aesthetic large model evaluation (semi-objective and semi-subjective), and EEG monitoring (subjective). As shown in Tab. \ref{tab:quality} \& \ref{datasets_vila}, the HealSoul dataset demonstrates higher aesthetic quality, while images generated by EmoMV exhibit superior image quality and aesthetic performance compared to HealSoul. Through emotional guidance via music, EmoMV enhances both the quality and aesthetic appeal of the images.

% As illustrated in \cite{zheng2015investigating}, $Alpha$ is Negatively related to good emotion, while $Beta$ and $Gamma$ wave are positively related to bad emotion. The $Mean$ of these wave represents the whole emotion level, while $Events$ denotes the sharp change of the wave which denotes the tension of the effect to people's emotion. We randomly selected 15 participants and conducted a random test of 100 image-music pairs with synchronized presentation, while monitoring EEG brainwave responses (as shown in Tab. \ref{tab:eeg}). The results indicate that EmoMV shows superior performance with more frequent change of good emotion and less of bad emotion, demonstrating better emotional expression and audiovisual consistency, thus proving EmoMV’s ability to generate images with consistent emotional dimensions.
As demonstrated in \cite{zheng2015investigating}, alpha waves are negatively correlated with positive emotions, whereas beta and gamma waves are positively correlated with good emotions. The ``mean" amplitude of these waves reflects the overall emotional state, while ``events" signify sharp fluctuations in brainwave activity, indicating emotional tension. To evaluate the effectiveness of EmoMV, we conducted a study with 15 participants who were exposed to 100 synchronized image-music pairs while their EEG brainwave responses were recorded (see Table \ref{tab:eeg}). The results indicate that EmoMV significantly enhances positive emotional responses and reduces negative emotional responses compared to baseline methods. Specifically, EmoMV-generated images elicited more frequent increases in positive emotions and fewer instances of negative emotions, demonstrating superior emotional expression and audiovisual consistency. These findings validate EmoMV’s ability to generate images with coherent and consistent emotional dimensions.

\begin{figure*}
    \centering
    \includegraphics[width=1\linewidth]{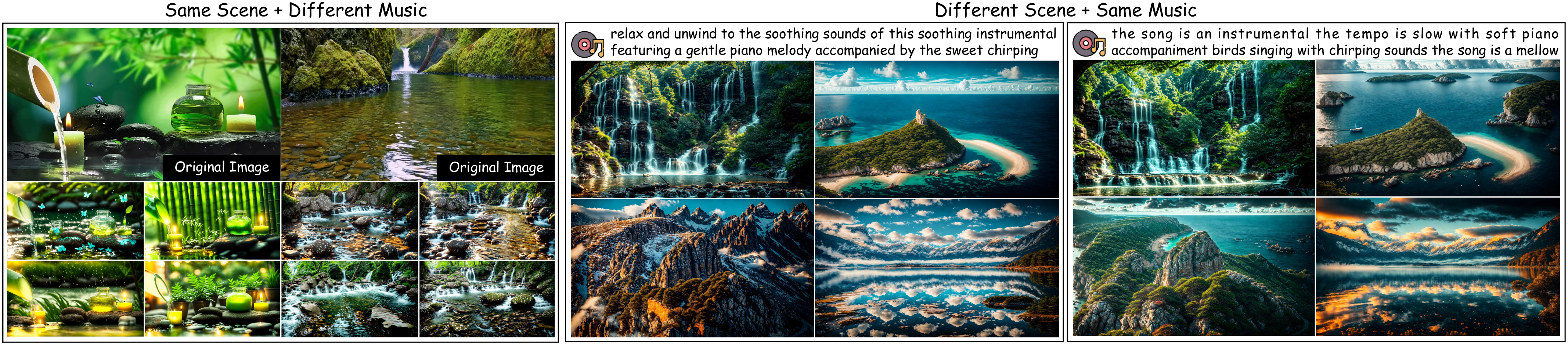}
    \caption{Case Study of EmoMV for Emotion-Driven Music-to-Image Generation.
    Left: Visual outputs generated from the same image with four different musical inputs, illustrating how variations in musical emotion (e.g., tempo, rhythm, mood) influence the aesthetic styling and emotional tone of the generated images using the EmoMV method.
    Right: Visual outputs generated from the same musical input but with four different scene images, demonstrating how varying visual contexts affect the interpretation of the music’s emotional cues and alter the visual aesthetic representation in the EmoMV framework.}
    \label{fig:case}
\end{figure*}

% \subsection{Analysis of the Quality of Images generated from EmoMV}
% The comparative analysis of the EmoMV-1k dataset (enhanced) and the HealSoul dataset (baseline) demonstrates that the proposed method of leveraging music emotion dimensions to modulate image illumination significantly improves overall image quality, as demonstrated in Tab. \ref{tab:quality}. The EmoMV-1k dataset exhibits notable advancements in several key metrics, including sharpness, contrast, and overall visual quality as measured by BRISQUE. Specifically, the mean sharpness value increased substantially, indicating enhanced detail preservation and improved image clarity. Contrast values also became more uniform, reflecting better tonal distribution and improved visual depth. Furthermore, the BRISQUE scores were markedly reduced, indicating a significant reduction in distortions and noise. In terms of colorfulness, the EmoMV-1k dataset displayed a more stable distribution, with a decreased standard deviation and fewer low-saturation or grayscale images, thereby improving the dataset’s visual consistency. While a small subset of images in the EmoMV-1k dataset still exhibits lower sharpness and quality, the overall improvements highlight the efficacy of the proposed approach. These results demonstrate that integrating music emotion dimensions into image enhancement processes provides a novel and effective means of improving image quality, offering substantial potential for further applications in cross-modal research.

\subsection{Ablation Study}
To evaluate the roles of the \textit{MVTA} and \textit{EAIR} modules in EmoMV, we conducted ablation studies. Removing the \textit{MVTA} module blocks the extraction of emotion-related and musical information, causing the EADL module to process only image inputs and pass Aes-Des directly to IC-Light. Similarly, excluding the \textit{EAIR} module by eliminating Aes-Des from EADG disrupts the Emotional Aesthetic Description Generation’s ability to transform information.
As shown in Table \ref{tab:quality_comparison_transposed}, EmoMV-1k achieves superior image quality, particularly in Sharpness and BRISQUE metrics. Figure \ref{fig:ablation} illustrates that both the ``w/o EAIR” and ``w/o MVTA” variants exhibit reduced consistency with the original image to varying degrees. Specifically, ``w/o MVTA” lacks narrative descriptions, creating a semantic gap, while ``w/o EAIR” produces chaotic images that fail to convey harmony or emotional coherence. These results demonstrate the essential contributions of both modules to maintaining image quality and emotional fidelity in EmoMV.
\begin{table}[htbp!]
\tabcolsep=0.1cm
\centering
\caption{Comparison of the mean value of Image Quality Metrics for \textit{w/o EAIR}, \textit{w/o MVTA}, and \textit{EmoMV-1k} Datasets. The best results are highlighted in \colorbox{lightred}{pink}.}
\begin{tabular}{lcccccc}
\toprule
\multirow{1}{*}{Method} & Sharpness                                                   & Contrast                                                & Colorfulness                                               & BRISQUE                                                   \\ \cline{2-5} 

\textit{w/o EAIR} 
    & $441$
    & \colorbox{lightred}{$1.00$} 
    & \colorbox{lightred}{$147.19$} 
    & $36.03$ \\ \midrule

\textit{w/o MVTA} 
    & $558.97$
    & $0.99$ 
    & $138.39$ 
    & $31.49$ \\ \midrule

\textit{EmoMV-1k} 
    & \colorbox{lightred}{$656.58$} 
    & $0.99$
    & $142.73$ 
    & \colorbox{lightred}{$28.07$} \\ \bottomrule
\end{tabular}
\label{tab:quality_comparison_transposed}
\end{table}

\subsection{Caes Study}

This case study examines how different musical emotions influence the aesthetic stylization of images generated by EmoMV. We present two scenarios: (1) the same image with different music, and (2) the same music with different images, as illustrated in Fig. \ref{fig:case}.
In the first scenario, four music tracks with varying emotional tones are applied to a single image. Despite using the same visual content, the resulting images exhibit distinct stylistic variations based on the emotional cues in the music. For example, gentle music creates warm tones, soft lighting, and smooth transitions, evoking a serene mood, while soothing music produces cooler tones and emphasizes natural elements, maintaining a tranquil atmosphere. Other tracks, with livelier or more melancholic moods, further highlight how music’s emotional properties can shape lighting, composition, and color tones, transforming the aesthetic style of the image.
The second scenario applies the same soothing music to four different scene images. Here, the music remains constant, but the generated visuals vary significantly depending on the scene. Despite the same soothing, tranquil music, the visual outcomes adapt to the context of each scene. In natural landscapes, the images maintain calm, cool tones that align with the peaceful nature of the music, while urban or abstract scenes tend to reflect subtler lighting adjustments and compositional changes that preserve the serene emotional tone of the music. These variations illustrate how EmoMV can modify visual outputs based on both the emotional input from music and the inherent characteristics of the visual scene, enhancing the emotional depth and aesthetic appeal of the images.
This case study demonstrates EmoMV’s flexibility in using music-driven emotional cues to create context-sensitive and emotion-driven visual stylizations, whether through varying music or diverse images.

\section{Conclusion}
In this paper, we present \textit{EmoMV}, a novel two-stage framework for emotion-driven image manipulation that integrates bottom-up and top-down cognitive strategies to construct emotional representations from music and visuals. We introduce the Mus-Vis Textual Alignment module to harmonize musical and visual descriptions with their emotional expressions and the Emotion-aware Aesthetic Image Refinement module to map these emotions into visual elements such as lighting and exposure, ensuring cohesive image harmony. Additionally, we develop a comprehensive multi-scale evaluation framework and validate EmoMV’s effectiveness through extensive experiments on a 38,000 music-image pair dataset, demonstrating its ability to translate musical emotions into compelling visuals and positively impact emotional well-being, particularly in applications like art therapy. These contributions advance the field of multimodal emotional integration, offering new avenues for creative industries and enhancing human-computer interaction.

\bibliographystyle{IEEEtran}
\bibliography{arXiv}

\end{document}